**Fausto Giunchiglia – DISI, University of Trento, Italy**
**Mayukh Bagchi – DISI, University of Trento, Italy**

# Object Recognition as Classification via Visual Properties

**Abstract:**
We base our work on the *teleosemantic modelling* of concepts as *abilities* implementing the distinct *functions* of *recognition* and *classification*. Accordingly, we model two types of concepts - *substance concepts* suited for object recognition exploiting visual properties, and *classification concepts* suited for classification of substance concepts exploiting linguistically grounded properties. The goal in this paper is to demonstrate that object recognition can be construed as classification via visual properties, as distinct from work in mainstream computer vision. Towards that, we present an object recognition process based on Ranganathan's four-phased faceted knowledge organization process, grounded in the teleosemantic distinctions of substance concept and classification concept. We also briefly introduce the ongoing project *MultiMedia UKC*, whose aim is to build an object recognition resource following our proposed process.

## 1.0 Introduction

Recognition, the acknowledgement of identity of senses across encounters or experiences, is fundamental to human perception. It is most prominently manifested in (human) vision, the fundamental faculty of which is to visually recognize our world in terms of objects, categories, and at a higher level, scenes and events. The inputs of visual recognition are then exploited to build concepts, which, though universally agreed as *mental representations*, differ widely in their modelling paradigm. The mainstream approach, termed *Descriptionism* in Millikan (2000), models concepts as *classes* described intensionally by their linguistically grounded properties, and drives phenomena such as (human) knowledge acquisition, reasoning and communication. An alternative approach termed *Teleosemantics* (Giunchiglia and Fumagalli 2016), inspired by the work of Ruth Garrett Millikan (Millikan 2000, Millikan 2004, Millikan 2005), was recently proposed which models concepts as *abilities* capable of implementing *functions* such as *visual recognition* and *classification*. This shift from modelling the means of static representation of the world to modelling the means of continual generation of such representations forms the basis for *visual semantics* (Giunchiglia, Erculiani and Passerini 2021), namely, the study of how concepts are generated from visual perception.

In the context of visual semantics, there are two *stratified* problems which are of particular concern for object recognition. The *first problem* is the many-to-many mapping between what is the case in the world, i.e., substances, and the visual input perceived from such substances, i.e., objects (termed *Sensory Gap* in Smeulders *et al.* 2000). The *second problem* results from the many-to-many mapping between the information conveyed by the visual input and its contextual interpretation by a user depending on purpose or objective to be achieved (termed *Semantic Gap* in Smeulders *et al.* 2000). These two problems, we argue, are a consequence of the central assumption prevailing in existing computer vision systems - that, in visual recognition, objects are hardly organized into classification hierarchies post visual perception. Furthermore, in the few studies they are organized as such (see, for instance, Marszalek and Schmid 2007; Deng *et al.* 2009), the classification is performed on linguistically grounded properties and not on visually perceived properties, thus generating the *many-to-many mapping*.

The goal of this paper, accordingly, is to model (visual) object recognition as a task of organizing objects into classification hierarchies based on visual properties via exploiting *visual genus-differentia* (Giunchiglia, Erculiani and Passerini 2021), with linguistically grounded properties only acting as vehicles for (formal) communication and reasoning. It does so by firstly introducing a novel process of object recognition through classification via visual properties, based on Ranganathan's four-phased knowledge organization process (Ranganathan 1967), grounded in the teleosemantics modelling of concepts as recognition and classification abilities respectively. Secondly, it introduces an ongoing project to develop a multimedia, multilingual lexical-semantic resource - named the *MultiMedia UKC* - which will solve the sensory and the semantic gap problem as noted before, with promising preliminary results.

The rest of the paper is organized as follows. Section 2.0 introduces the sensory gap and the semantic gap problem. In Section 3.0, we contextualize the teleosemantics notions of substance and classification concept in terms of the sensory and semantic gap problem. Section 4.0 details our *object recognition process* which solves the sensory and the semantic gap problem. In Section 5.0, we provide a birds-eye view of the ongoing project on the *MultiMedia UKC*. Section 6.0 concludes the paper.

## 2.0 The Sensory Gap and the Semantic Gap Problem

Object recognition, that we model here as the *ability* to learn, classify and (re-)identify a diverse array of visual objects over encounters, is an outstanding challenge for mainstream computer vision systems. It is particularly difficult due to two *stratified impediments*. Firstly, the process of perception of an object is complicated by the interference of *confounding variables* such as clutter and occlusion which are ubiquitous in mainly, but not only, *open world settings* (Bendale and Boult 2015) by which we mean settings which mirror human visual perception and recognition. Secondly, alongside perception, the fact that the process of (incremental) recognition and subsequent (system) representation of the object is *not a single, general purpose process* (Logothetis and Sheinberg 1996) but *interpretative* in nature, involving, as has been elaborately established in neuroscience (Kandel *et al.* 2000; Martin and Chao 2001), multiple representations of what is perceived stemming from (combinations of) multiple viewpoints or aspects. The two impediments above afflicting object recognition have been crystallized into two *stratified* problems (Smeulders *et al.* 2000) - the Sensory Gap Problem and the Semantic Gap Problem.

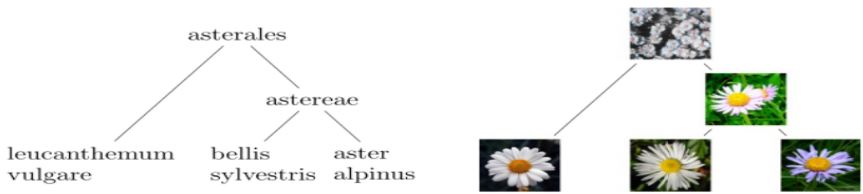

**Figure [1]: Classification of *Daisies* - in words (left); in pictures (right)**

Firstly, the *Sensory Gap Problem* results from:

"the gap between the object in the world and the information in a (computational) description derived from a recording of that scene" (Smeulders et al. 2000, 1352).

Let us illustrate the problem with an example scenario of a florist utilizing a *smartglass* to attempt to recognize *daisies* in an open setting such as, for example, in a commercial nursery (see Figure [1] taken from Giunchiglia, Erculiani and Passerini 2021). Firstly, the object *daisy* can only be *partially perceived* by the system over encounters, due to several possible factors such as, for instance, the differences in the orientation of the daisy during perception, possible clutter and occlusion with respect to the visual field. This results in the persisting *sensory gap* between the *object* daisy and the *information derived* about it from (perceptual) recording of the scene by the system, in other words, the *visual data*. The *Semantic Gap Problem* is manifested on top of the sensory gap due to:

> "the lack of coincidence between the information that one can extract from the visual data and the interpretation that the same data have for a user in a given situation" (Smeulders *et al.* 2000, 1353).

Given the premise of partial perception over encounters, the fact remains that the daisy, depending on the context, focus or purpose, can be recognized and subsequently represented in the system as a *daisy* or as a *asterales* or as a *leucanthemum vulagre* or even as *daisy#123*, a daisy which is studied as a part of an experiment. Further, the recognition and representation of a daisy as *asterales* or a *leucanthemum vulagre* can be the result of (partial) perceptions with respect to the same or different daisies. This results in the *semantic gap* - the *lack of coincidence* between the information conveyed by the *visual data* and its contextual interpretation for a user.

The above discussion on sensory and semantic gap merits three crucial observations. Firstly, the impossibility of avoiding the confounding variables in real world visual recognition due to open world settings. In fact, the very notion of open world settings is grounded in what are known in computer vision as *broad domains* (Smeulders *et al.* 2000), namely visual domains the objects of recognition of which are highly *visually polysemic* and hence, potentially exhibit *unlimited and unpredictable variability* rendering their semantic conveyance partial. Secondly, the lack of coincidence between the *visual data* and its contextual interpretation for a user entails the assumption of an *egocentric point-of-view* (Smith and Slone 2017; Erculiani, Giunchiglia and Passerini 2020), the notion that the interpretation is bound to the inherent purpose of a single individual, and hence, can't be a single, general purpose process. Thirdly, in light of the above two observations, the issue that visual recognition of the two types - recognizing *categories* and recognizing *instances* of those categories, until now, has been *tackled distinctly* via dedicated algorithms for each type (Grauman and Leibe 2011).

In computer vision, the prevailing approach has been *content-based* object recognition which, ideally, refers to recognition via exploiting *low-level content* (such as color, shape, texture etc.) of the object rather than *high-level metadata* (such as labels, description etc.) of its multimedia (such as image, video etc.) representation. It is, however, clear from studies such as (Marszalek and Schmid 2007; Porello, Cristani and Ferrario 2013) that the focus has been on training *learning algorithms* for categorizing output images which are, subsequently, attached to nodes from a linguistically grounded classification hierarchy (such as WordNet). There are *three clear shortcomings* in the above mainstream approaches. Firstly, there is no systematic mechanism in any of the above approaches to accommodate and formalize the sensory gap. Secondly, all of the above approaches employ *a priori knowledge* (such as WordNet, or specific ontologies) to enforce classification of multimedia representations, which is in *direct contrast* to the notion of *content-based* recognition due to the fact that they construct the classification (subsumption) hierarchy based on *linguistically defined properties* (high-level content) and not, *ab initio,* from *visual properties* via exploiting *visual genus-differentia* (low-level content). Thirdly, as a consequence of the above contrast, there is no object recognition process for solving the semantic gap which is, in other words, the misalignment between the visual subsumption hierarchy built via exploiting *visual properties* which is *egocentric* in nature (right hierarchy of Figure [1]), and the linguistically grounded subsumption hierarchy (left hierarchy of Figure [1]).

## 3.0 Substance Concept & Classification Concept

Following recent results in *Teleosemantics* (Macdonald and Papineau 2006), formerly known as *Biosemantics* (Millikan 1989), we propose two postulates. Firstly, we posit that the world is comprised of *substances* (Millikan 2000) which are:

"those things about which you can learn from one encounter something of what to expect on other encounters, where this is no accident but the result of a real connection" (Millikan 2000, 15).

Notice that we restrict ourselves to substances which can only be visually perceived (i.e. *objects*), and the role of substances beyond visual perception is irrelevant for us. Secondly, we model concepts as (mental) *abilities* (as in Giunchiglia and Fumagalli 2016) implementing suitable (*etiological*) *functions* which must be understood as *'intended for'* or *'devised for'* a specific purpose (Neander 1991). Such a proposition cumulatively enables us to distinguish between the two abilities which drive object recognition - *visual recognition* and *classification*, and accordingly, facilitates the introduction of two novel type of concepts, namely, *substance concepts* which are focused on (continual) *visual recognition* of substances, and *classification concepts* geared towards *description and classification* of substances post visual recognition (Millikan 2000; Giunchiglia and Fumagalli 2016; Fumagalli, Bella and Giunchiglia 2019).

As from the definition of substances, their recognition happens only *indirectly* via *sets of encounters*, with encounters being events in which substances *partially expose* themselves to visual object recognition systems over *space-time*. Substance concepts are *abilities* (in our case, of visual recognition systems) which implement the *function* of visually recognizing substances over encounters in an *egocentric* fashion. The recognition process of substance concepts is *uniform* as to whether the substance is an *individual* (such as *daisy#123*) or a *real kind* (such as *daisy*), with the exact granularity of visual recognition being left to the purpose or objective of the user of the visual recognition system. Further, the key to continual recognition of a substance as a substance concept is grounded in its *causal factor*, which are a set of inner characteristics of a substance responsible for its *broad invariance* across encounters (such as *homeostasis* in living beings). These inner characteristics are manifested as perceivable *visual properties* (low-level content), which, in turn, facilitates us to (incrementally) model the substance as *visual objects*, where visual objects are sequences of similar *visual frames* (Erculiani, Giunchiglia and Passerini 2020; Giunchiglia, Erculiani and Passerini 2021). Concretely, in the right hierarchy of Figure (1), in each set of encounters with the daisies under observation, the smartglass perceives only a *partial view* of them, and accordingly, generates substance concepts of daisies at different levels of abstraction such as *asterales*, *astereae*, *leucanthemum vulagre*, *aster alpinus*.

Differently from substance concepts, classification concepts are *abilities* which implement the function of (continually) *describing and classifying* substance concepts post their visual recognition. This is done via the following *continual* mechanism. Firstly, classification concepts annotate the (incoming) substance concepts with a unique, linguistically-grounded *word sense* (Miller 1995). Secondly, such annotated concepts are organized into (lexical-semantic) classification hierarchies following the intensional paradigm of *genus-differentia* (Parry and Hacker 1991), wherein, *genus* is an intensional definition marked by a set of properties which are common across distinct objects, and *differentia*, instead, are the set of properties facilitating discrimination amongst objects with the same *genus*. A fitting example, for instance, would be the *differentia* for *astereae* and *leucanthemum vulagre* which are daisies with the same *genus*, namely, *asterales* (left hierarchy in Figure [1]).

There are four highlights which are relevant for the above teleosemantics distinction between substance concepts and classification concepts. Firstly, the fact that the very notion of modelling concepts as *abilities* is in perfect sync with the procedural nature of visual object recognition which, by itself, is an *ability* (see Section 2.0). Secondly, the modelling of visual objects via visual frames guarantees the facility of computing visual (dis)similarity among what is perceived. Thirdly, the pervasive *obfuscation* in mainstream vision systems between the representationally distinct concepts of *recognition* and *classification*, the former being modelled in terms of temporal sequences of visual frames (*visual data*; such as 2D photos or 3D videos), and the latter modelled linguistically in terms of glosses enunciating genus and differentia. Finally, and perhaps the most important, the distinct *ontological nature* of objects which underlie the distinction between the two types of concepts (Giunchiglia, Erculiani and Passerini 2021). When modelled as substance concepts and classification concepts, objects are construed, respectively, as *perdurants* - entities which are only *partially present* at any moment in time, and thus have *temporal parts*, and *endurants* - entities which are *wholly present* at any moment in time (Gangemi *et al.* 2002).

There are, however, two crucial issues which are still outstanding. The first issue being that the very definition and the process by which substance concepts are generated ensure the *necessary* but *not sufficient* (incremental) resolution of the *sensory gap*. It does provide a definition and process by which substance concepts (*visual data*) are generated from objects (*necessary condition*), but it doesn't provide a computational mechanism (such as memory) in which identity of the substance concepts can persist and be incrementally refined (*sufficient condition*). The second issue being the fact that, even though the two types of concepts are equipped to accommodate the *many-to-many mapping* between the object, it's visual data (representation) and the contextual interpretation of it by the user, there is *no concrete process yet* to concretely effectuate it's resolution (and hence, the *semantic gap* persists). This is exactly where the work in dynamic knowledge organization by Ranganathan (1967) provides a major leap.

**4.0 The Object Recognition Process**

| Knowledge Organization Process | Object Recognition Process |
|---|---|
| Pre-Idea Stage | Substance Concept Recognition |
| Idea Plane | Substance Concept Hierarchy Construction |
| Verbal Plane | Linguistically-labelling Substance Concept Hierarchy |
| Notational Plane | Classification Concept Hierarchy Formalization |

**Table [1]: The Object Recognition Process**

Independently from the teleosemantics modelling of concepts as abilities, concepts in mainstream faceted knowledge organization (Broughton 2006; Giunchiglia, Dutta and Maltese 2014) are organized according to the *analytico-synthetic* paradigm (Ranganathan 1967; Ranganathan 1989). We are particularly interested in the *stratified process* proposed by Ranganathan (1967) which unifies the perceptual recognition of concepts with their (lexical-semantic) description and classification. The process is implemented in four phases - *Pre-Idea Stage*, *Idea Plane*, *Verbal Plane* and *Notational Plane*, with the motivation behind such stratification being the *understanding and exploitation* of each phase in a characteristically autonomous yet functionally linked fashion. A key highlight of the process is in how it provides an *innate mechanism* conjoining substances, substance concepts and classification concepts, resulting in the object recognition process (Table [1]).

The *Pre-Idea Stage* is *causal* by nature, and is focused on continual *cognitive grounding* of concepts, whereby, they are (visually) perceived, recognized and finally (mentally) represented in a memory. Firstly, perception, which is the *reference of a percept to its entity-correlate outside the mind* (Ranganathan 1967), is incrementally facilitated by *pure percepts*, which are meaningful impressions of an entity (object) generated by a single primary sense (vision) and deposited in the memory. This process of perception, over several sets of encounters, results in *compound percepts*, which are agglomerated (meaningful) impressions - *concepts* - formed via association of several pure percepts in quick succession, deposited in the memory. Such evolving assimilation of newly perceived percepts with pre-existing concepts in the memory is referred to as *apperception*, with the continuously evolving memory being the *apperception mass* (Ranganathan 1967). For example, the smartglass perceives a daisy (*object*) from a horizontal view on one set of encounters (*pure percept*). During the next set of encounters, it recognizes the same daisy, but this time from a vertical view (a different *pure percept*). The system associates the two pure percepts to form the compound percept *daisy*, in other words the concept *daisy*, and deposits it in the *apperception mass* for (possible) further evolution. Clearly, analyzed from the perspective of teleosemantics (Table 1), the concepts recognized and represented in this phase correspond directly to *substance concepts* with the *key* addition of apperception mass, which is what we computationally model as *cumulative memory* (see Giunchiglia, Erculiani and Passerini 2021).

The *Idea Plane*, instead, is focused on the organization of concepts recognized and agglomerated in the Pre-Idea Stage into a 'perceptual' classification hierarchy, such as, in our case, *visual subsumption hierarchies* (right hand side of Figure [1]) constructed by organizing substance concepts as a function of their *visual properties* via exploiting *visual genus and differentia* (Giunchiglia, Erculiani and Passerini 2021). It is, indeed, a phase which serves as the *map and foundation* (Satija 2017) for the classification hierarchy as they are not constructed *based on intuition* but informed by an exhaustive set of *tried and tested* canons for designing classification systems. The key observation here, however, is the fact that the set of canons for characteristics, succession of characteristics, arrays and chains originally proposed by Ranganathan (1967) holds majorly, but not entirely, for visual genus and differentia (detailed in Giunchiglia and Bagchi 2021). A prominent example being the set of principles of *helpful sequence* which, being designed for user-centred shelf arrangement in libraries, is irrelevant for our work in object recognition.

While the Idea Plane is focused on the classification of substance concepts by exploiting their visual properties, the *Verbal Plane* and the *Notational Plane* focuses on linguistic rendering of the perceptual (visual) hierarchies generated by the Idea Plane, and thus, *converting substance concepts into classification concepts*. This is crucial because, while visual classification is apt for *recognition*, it is not suitable for (formal) communication and reasoning, for which we require *language* (Fumagalli, Bella and Giunchiglia 2019). The focus of the *Verbal Plane* is to exploit linguistic labels, i.e. words from an appropriate *object language* (Ranganathan 1967) such as natural languages or domain namespaces to annotate the visual hierarchy (of photos) while respecting *Genus-Differentia*. It is worth noting that these linguistically-labelled hierarchies are not only distinct for each object language due to *multilingual lexical gaps* (as studied in Giunchiglia, Batsuren and Bella 2017 and Giunchiglia, Batsuren and Freihat 2018), but also highly *egocentric* depending on one single individual's purpose. Finally, the *Notational Plane* formally identifies each concept in the language-annotated hierarchy with unique identifiers, thus mitigating any remaining linguistic phenomena such as *polysemy, homonymy* and *synonymy* (Giunchiglia, Batsuren and Bella 2017), and, in the process, converts substance concepts into formal, uniquely identified classification concepts. The canons for the above two phases are detailed in Giunchiglia and Bagchi (2021).

There are five significant takeaways from the above discussion. Firstly, the central role of perception and visual properties, and, thereby substance concepts, in the first two phases, *viz*. Pre-Idea Stage and Idea Plane, which cumulatively decides the organization of concepts. Secondly, the *Wall Picture-Principle* (Ranganathan 1967), namely the notion that idea precedes language provides justification for the teleosemantics distinction between substance concept and classification concept. Thirdly, the introduction of cumulative memory solves the *sensory gap* problem in a necessary and sufficient manner. Fourthly, and most importantly, the process *enforces* a *one-to-one mapping* between substance concepts and classification concepts, and thus solves the long-standing *semantic gap* problem. Finally, the fact that the above methodology is fully *egocentric* in the sense that the incremental design of the hierarchy completely depends upon the purpose or objective of the user, the latter being guaranteed by the *Law of Local Variation* (Ranganathan 1967).

## 5.0 The MultiMedia UKC

The rise of multimedia, in the form of images (such as in Instagram, Pinterest), videos (such as in YouTube), and even in e-commerce catalogs (such as Amazon), fuelled the development of several *benchmark* image databases in Computer Vision. Amongst them, ImageNet (Deng *et al*. 2009) is by far the largest image database employed to train deep neural networks for developing content-based image understanding algorithms. It has been designed by annotating each synset of the noun hierarchy of the Princeton WordNet (Miller 1995) with, on average, about 600 photos curated from the crowdsourcing platform Amazon Mechanical Turk. The key observation, however, is the fact that by its very design, ImageNet suffers from both the sensory gap and the semantic gap. Since ImageNet is constructed by populating a linguistic hierarchy with photos, it is by design a classification concept hierarchy and the substance concepts are encoded via the experience of the people who have classified the photos inside the WordNet hierarchy. But the coincidence between this and the images classified is far from obvious, and almost never the case, this being exactly the reason why the sensory and the semantic gap are pervasive.

The aim of the project introduced here, named *MultiMedia Universal Knowledge Core*, or *MultiMedia UKC* in short, is to build a resource similar to ImageNet but with the key distinction of being incrementally developed following our proposed object recognition process. The ultimate goal is to build a sensory and semantic gap free ImageNet- like hierarchy. We co-opt the existing *Universal Knowledge Core* (UKC; Giunchiglia, Batsuren and Bella 2017, Giunchiglia, Batsuren and Freihat 2018), a multilingual lexical-semantic resource of over one thousand languages and over one hundred thousand classification concepts, as our starting base. The advantages of exploiting the UKC is in its innate organizational similarity to Ranganathan's knowledge organization process, and in its large-scale multilinguality which opens up the possibility for multimedia, multilingual lexical-semantic hierarchies. Concretely, following our proposed object recognition process, the ongoing development of *MultiMedia UKC* proceeds as follows-

1. *Pre-Idea Stage*, wherein substance concept recognition takes place via extraction of visual objects from media, following the techniques elucidated in Erculiani, Giunchiglia and Passerini (2020) and Giunchiglia, Erculiani and Passerini (2021).
2. *Idea Plane*, wherein a visual subsumption hierarchy of visual objects is constructed by organizing them as functions of their *visual properties*. As observed in Giunchiglia, Erculiani and Passerini (2021), human supervision is crucial in this phase for fixing of *genus-differentia* respecting the canons detailed in Giunchiglia and Bagchi (2021).
3. *Verbal Plane*, wherein each node in the visual subsumption hierarchy is labelled with appropriate words, synsets and glosses from an object language. In this phase, the *Language Core* of the UKC, with over one thousand natural and domain languages, is heavily exploited.
4. *Notational Plane*, wherein the linguistically-labelled concepts are annotated with unique, *alinguistic* identifiers generating fully formal classification concepts. In this phase, the *Concept Core* of the UKC, with over one hundred thousand uniquely identified concepts, is heavily exploited.

Some early studies, as published in Giunchiglia, Erculiani and Passerini (2021) have already been produced. At the moment, a general algorithm whose aim is to automate the generation of the object hierarchy is under development.

**6.0 Conclusion**

We showed, based on recent results in teleosemantics and computer vision, that visual object recognition can indeed be seen as recognition and classification of objects based on their visual properties, with linguistically grounded properties chiefly employed for formal representation and reasoning. The research reported here are first steps towards full-fledged development of the *MultiMedia UKC*, an overview of which was also provided.

**Acknowledgement**

The research conducted by the authors has received funding from the "DELPhi-DiscovEring Life Patterns" project funded by the MIUR Progetti di Ricerca di Rilevante Interesse Nazionale (PRIN) 2017 – DD n. 1062 del 31.05.2019.